# An Information Theoretic Approach to Sample Acquisition and Perception in Planetary Robotics

Garrett Fleetwood
Space and Terrestrial Robotics Exploration Laboratory
School of Energy, Matter and Transport Engineering
Arizona State University
Tempe, Arizona, United States
gfleetwo@asu.edu

Jekan Thangavelautham
Space and Terrestrial Robotics Exploration Laboratory
School of Earth and Space Exploration
Arizona State University
Tempe, Arizona, United States
jekan@asu.edu

*Abstract*—An important and emerging component of planetary exploration is sample retrieval and return to Earth. Obtaining and analyzing rock samples can provide unprecedented insight into the geology, geo-history and prospects for finding past life and water. Current methods of exploration rely on mission scientists to identify objects of interests and this presents major operational challenges. Finding objects of interests will require systematic and efficient methods to quickly and correctly evaluate the importance of hundreds if not thousands of samples so that the most interesting are saved for further analysis by the mission scientists. In this paper, we propose an automated information theoretic approach to identify shapes of interests using a library of predefined interesting shapes. These predefined shapes maybe human input or samples that are then extrapolated by the shape matching system using the Superformula to judge the importance of newly obtained objects. Shape samples are matched to a library of shapes using the eigenfaces approach enabling categorization and prioritization of the sample. The approach shows robustness to simulated sensor noise of up to 20%. The effect of shape parameters and rotational angle on shape matching accuracy has been analyzed. The approach shows significant promise and efforts are underway in testing the algorithm with real rock samples.

## I. INTRODUCTION

Planetary exploration has benefitted from rapid advancement and miniaturization of electronics, sensors and actuators. Advancements have been made in autonomous control systems, particularly in simultaneous, localization and mapping (SLAM), obstacle avoidance, low-level planning and vision. However, planetary robotics typically requires humans in the loop, particularly, teams of missions scientists to guide their every move as the missions are science focused. The long latencies of 9 to 19 minutes to Mars, the relatively low communication bandwidth all slow the tempo of operations. These challenges will reduce the effectiveness and efficiency of robotic sample retrieval and sample return missions. According to the Planetary Science Decadal Survey [1], sample retrieval and sample return are identified as upcoming major goals in planetary science. They are the focus of upcoming Mars 2020 rover and future missions. Identification and return of the right sample can provide significant insight into a planet's origins including its geo-history and origin of life. A good example is the Mars meteorite NWA 7034, "Black Beauty" found in the Western Sahara (Fig. 1) [2]. The sample has provided planetary scientist insight into Mars' geo-history over its first 1.6 billion years.

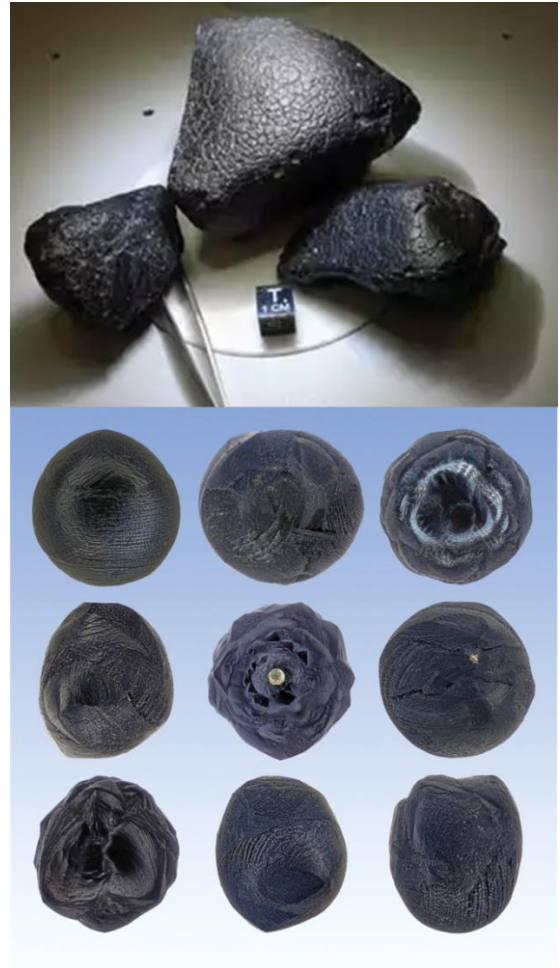

Fig. 1. Extraterresterial samples, including Mars Meteorite NWA 7034 [2] (top) and space dust magnified 3,000 times [3] (bottom) can give new insight into the geological history of early planetary bodies.

Finding such samples on the Martian surface will be a daunting challenge. If such processes can be partly or fully automated, it would enable planetary robotics systems to sift through more samples efficiently, prioritize and identify the most scientific interesting samples for further examination by the human science team. Beyond planetary exploration, this approach has relevance in identifying extra-terrestrial samples here on Earth. This includes stardust and particulate deposit from our neighboring planets, moons and asteroids (Fig. 1) [3]. Recent work has shown that these particles are everywhere on earth and at present requires a trained eye to distinguish between earth based dust particles [3].

In this paper, we propose an autonomous robotic control system to examine rock samples and categorize them by finding nearest match for them from a library of interesting rock samples. The matching process uses a Gaussian sensor noise model to generate probability of sample points used to describe a sample. The probabilities of points are used to determine the information content of the sample. Using the total information content, the shape is described using a linear set of eigenfaces of already identified library of "interesting" shapes [4].

By composing the identified shape as a set of eigenfaces, it is then possible to determine the best match of the sample to a library of interesting shapes. The shapes are discounted if it fails to show closest match beyond an information theoretic threshold. The library of interesting shapes is generated using the Superformula developed by botanist Johan Gielis [5]. Alternately, the library maybe generated using human input and samples. The Superformula has been shown to be used to describe natural objects including rocks, plants, trees and animals. The algorithm can generate both symmetrical and non-symmetrical shapes. Hence, the approach permits matching against a diverse number of interesting objects without having to have a library of 3D shape models. Our approach allows for extrapolation of the library of relevant interesting shapes based on a given set of samples.

Our paper is organized as follows: Section II presents related work, followed by presentation of the algorithm in Section III, analysis of the algorithm in Section IV, followed by discussion in Section V and conclusions in Section VI.

## II. BACKGROUND AND RELATED WORK

There exists a significant number of exploration planning algorithms capable of navigating and mapping previously unknown environments [6-9]. However, these algorithms are most often designed to maximize the efficiency of mapping an area without incorporating the possible value of other factors like object analysis (for sampling) and evaluation.

One of the more rigidly defined challenges of this line of research is the definition of the specific metric by which interest, value and cost are to be judged [9]. The quantification of interest is a function of the definition of interest. Similarly, the quantification of cost must follow logically from a well-defined framework. There must also be a method of relating the relative importance of these two qualities. Fortunately, while the method of comparing the costs and benefits is fundamentally subjective and contextual, the costs themselves may be far more easily quantifiable.

An important challenge associated with this project is identifying a point of interest and classifying it as interesting. There are a plethora of different variables that could be assessed when considering interest. Temperature, shape, and color are just a few of the possibilities [9]. Even after defining this of these variables to consider, categorizing them or quantifying them in terms of interest is not a straight forward task. Some have simple and immediately apparent approaches. For example, it may be easy to determine that a region is unexpectedly hot and therefore worthy of investigation because its temperature differs significantly from the surrounding environment. Other characteristics, however, are much more difficult to analyze. Shape in particular presents some unique challenges. First, shapes are, in a sense, subjective. While there is an established mathematical definition of shape, it's only really effective for categorizing identical shapes [10]. There are statistical approaches, but they tend to presume a foreknowledge of feature correspondence and/or discard important shape information [10]. Other approaches using trained neural networks to identify or organize shapes [11],[12],[13],[14]. Consequently, specific mathematical definitions of shape are insufficient for performing comparisons based on sensory similarity.

After an object has been classified, another problem emerges: how to assign a metric that defines the interest value of that object. This is a separate but related challenge to defining a metric of interest. Each prototypical object or class of object can be assigned an interest value but it is highly unlikely that any object discovered in an environment will perfectly match one of these categories. As a result there must be some method of incorporating the quality of the categorization into the assigned interest value. Much like matching itself, the key difficulty of this task is assigning a numeric value to this quality. From a human perspective, it is fairly easy to draw these distinctions qualitatively. For example, it is fairly easy to say that an ellipse is more like a circle than it is like a square. This comparison does not, however, present a metric that defines how much more similar to a circle an ellipse is than it is to a square.

From the standpoint of the categorization of real objects, the answer to these questions depends on the task. If it is a task in which the object to be identified can take either curved or straight forms, or forms with convex and concave curves, then these naturally fall into the same category. The inverse is true if each shape exists in a separate category. Similarly, the importance of closeness of match can vary with specifications. If the objects being sought have a wide variation then matches less similar to the prototypical shape are still valuable, while objects that have little variation should closely hew to the definition of the category. The challenge that derives from this, then, is to develop a system that can be adjusted to match a wide variety of shapes to a category depending on the needs of the situation.

In this work, we set out to develop a system of categorizing the interest value of detected features. This system is intended to assign interest values to categories of detected features based on an information-theoretically based shape-matching technique that will provide the most likely fit for the detected object and a natural modifier that can change the weight given to the shape based on the quality of the match to the figure.

For the task of identifying interesting shapes, we first develop techniques to assemble a library of interesting shapes. One promising method for generation of these interesting shapes is using the Superformula developed by Gielis [5]. It is a simple formula that allows for the generation of complex, asymmetric shapes that can mimic the shapes of a wide variety of natural, organic, and man-made objects using only a small handful of inputs. The simplest form of the equation is:

$$r(\theta) = \left( \left| \frac{\cos(\frac{m_1 * \theta}{4})}{a} \right|^{(n_2)} + \left| \frac{\sin(\frac{m_2 * \theta}{4})}{b} \right|^{(n_3)} \right)^{\left(\frac{-1}{n_1}\right)}$$

Its coefficients can be modified to produce different shapes taking a number of very different forms. These include the geometric, organic looking, as well as strange, asymmetric shapes that are difficult to describe. Some examples of three-dimensional shapes generated using the Superformula is shown in Fig. 2. The Superformula has found a niche in gaming. In particular, games such as "No Man's Sky" are thought to use Superformula-like algorithms to dynamically generate natural/near natural looking objects in a computer generated universe.

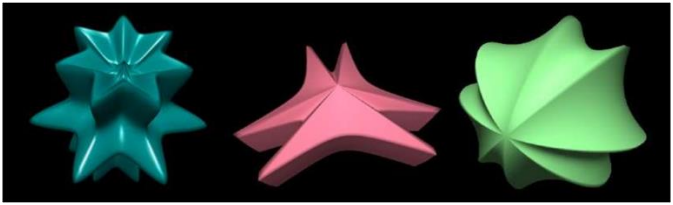

Fig. 2. Example shapes of generated using the Superformula [5].

This computer generated universe is expansive, yet has significant microscopic detail, with rocks, trees, animals and other object described using a Superformula like algorithm (Fig. 3).

### III. ALGORITHM

In this section, we describe the shape matching algorithm used to detect and classify shapes of objects founds against a library of desirable shapes. Fig. 4 shows the major steps in the algorithm. This shape matching process starts with detection of a rock sample and conversion of the outlines of the sample into polar coordinates. Once the base shape generation process is completed, the information measuring is started. The first component of this process is the generation of a probability distribution. For this work, a normal distribution was selected because it is the simplest kind of probability distribution to work with and not an unreasonable approximation of the error profile of many sensors.

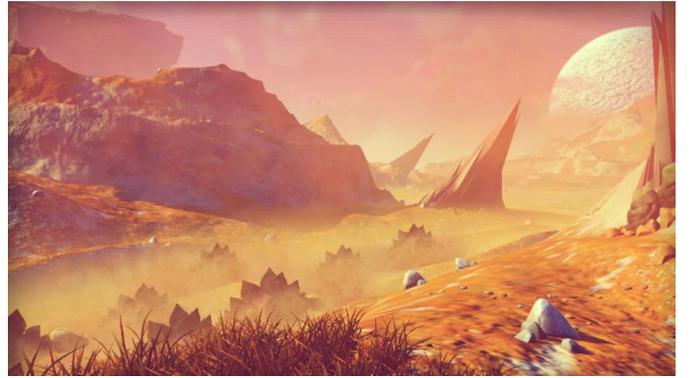

Fig. 3. Screenshot from the video game, "No Man's Sky" released in 2016. This space exploration game consists of a procedurally generated artificial universe with objects including rocks, plants and animals generated using Superformula-like algorithms.

The default value of standard deviation for this distribution was chosen to be 0.1. Using the calculated normal distribution, the process generates the probability components. The first is created by generating a series of one-dimensional rays of length 1/unit corresponding to every angle used in the re-generation of the shapes. These rays are then divided into sections, with each section being the length of a unit of resolution. The cumulative distribution function of the previously generated normal distribution is then used to calculate the probability of an obstruction existing with each section. This probability is calculated based on the assumption that the points should exist on a circle of radius $0.5\sqrt{2}$ centered in the middle of the analysis map.

These probability values are then converted into information units using Shannon's source information equation:

$$H(r, \theta) = p(r, \theta) * \log_2 \left( \frac{1}{p(r, \theta)} \right)$$

where p(r,θ) is the probability of an obstruction existing in section *r* of ray *θ*. While any information unit should be functional for this process, Shannon's bits were selected for the purposes of this work.

After this process is complete, a separate set of probabilities is generated. These values are to be used to modify the probability vectors generated in the previous process. These probabilities are generated using the same rays as the first set of probability values. However, instead of calculating their probability value relative to the radius of some circle, the points are calculated from a normal distribution centered on the radii of one of the generated shapes. The information values calculated in the previous section are then multiplied by this value to produce a unique projection of each shape.

Once these values have been calculated, they are plotted onto a grid map. This is the final result that is processed through the matching algorithm. The following examples are a few of these shapes and the Superformula shapes that they have been generated from. The logic behind this process is fairly simple. It is assumed that if there is an obstruction point than it is part of some larger object. The sensor scanning this object will have a corresponding error. The equation used assumes that all obstructions will take the shape of a circle and calculates the information value of finding a point anywhere on the map based on that circle and the sensor standard deviation. The choice of a circle is arbitrary and any shape could be substituted, a circle was selected primarily for the sake of simplicity. The specific equation used is the one Shannon used to determine the information value of an information source.

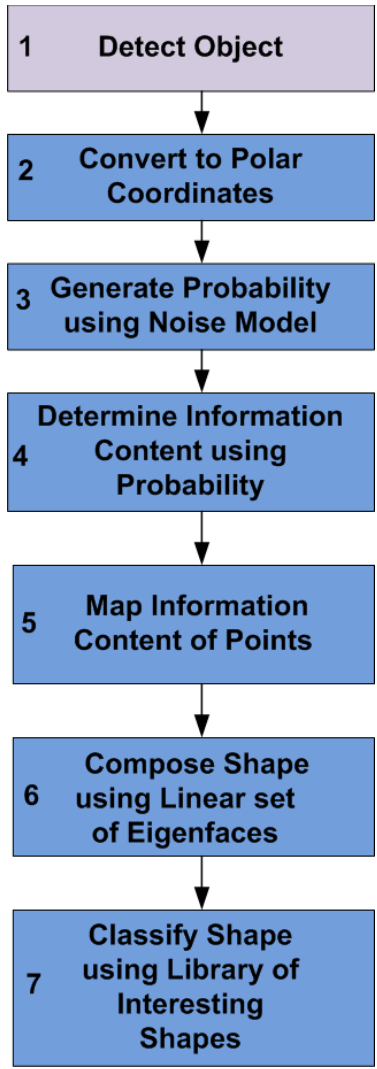

Fig. 4. The shape matching algorithm.

The values are distributed instead of summed. Initially it seemed that generating information values directly from the shape would not work. Realizing that the eigenfaces algorithm is an image matching technique, a work-around was created implementing the circle based technique with the modifiers. The logic of it, relating back to information theory, is that this will correspond to a measure of the information gained by knowing a point exists at a location relative to both the assumed circle and the shape it is theorized to be a component of.

A second variant of the technique makes the comparison not against a circle but against a defined shape. Instead of using the probability of an obstruction existing at a point relative if the object is assumed to be a circle, it is the probability that an obstruction exists at that point if the object is scanned is one of the Superformula generated shapes.

Otherwise, it is generated using the elements as the implemented equation. The logic of this process with relation to Information Theory is much clearer. Using this technique, if one were to trace a shape in a circle around the center of the map and collect one of the values at each angle it would be a representation of the amount of information gained by discovering that that was the actual shape rather than the Superformula shape it is assumed to be. Naturally, as this value decreases, the actual shape becomes increasingly similar to the Superformula shape it is being matched against. The following shape is generated using this technique (Fig. 5).

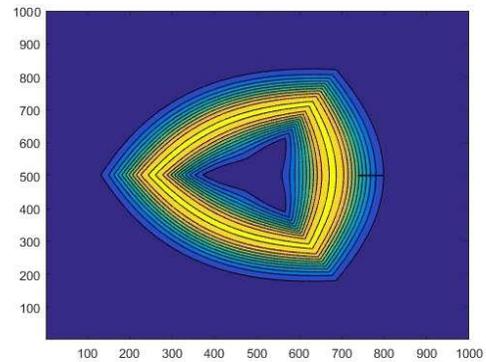

Fig. 5. Family of rounded-triangles generated using the Superformula.

The lighter areas represent higher values and the darker ones represent lower values. The high values are generally on the order of $10^{-4}$ and the lowest non-zero values are usually on the order of $10^{-12}$. These values can vary by the shape used.

*A. Matching*

The next important step in the process is to match the given shape against a library of interesting shapes. We first generate a set of eigenfaces of the information weighted set of points of each shape according to the process outlined in [4]. Examples of the information weighted points of sample shapes are shown in Fig. 6. Using the eigenfaces algorithm it is then possible to decompose the shape to classify into a linear combination of eigenfaces. Here we describe how we develop the eigenfaces in further detail. The eigenfaces algorithm takes as its initial input a set of shape images, these are known as the training shapes (shapes of interest). These images are input as M × N grids. These images need not be square, but all images must be of the same dimension (M × N). The pixels in these images are then reduced to grayscale values and the grid itself is converted into a vector of length M·N.

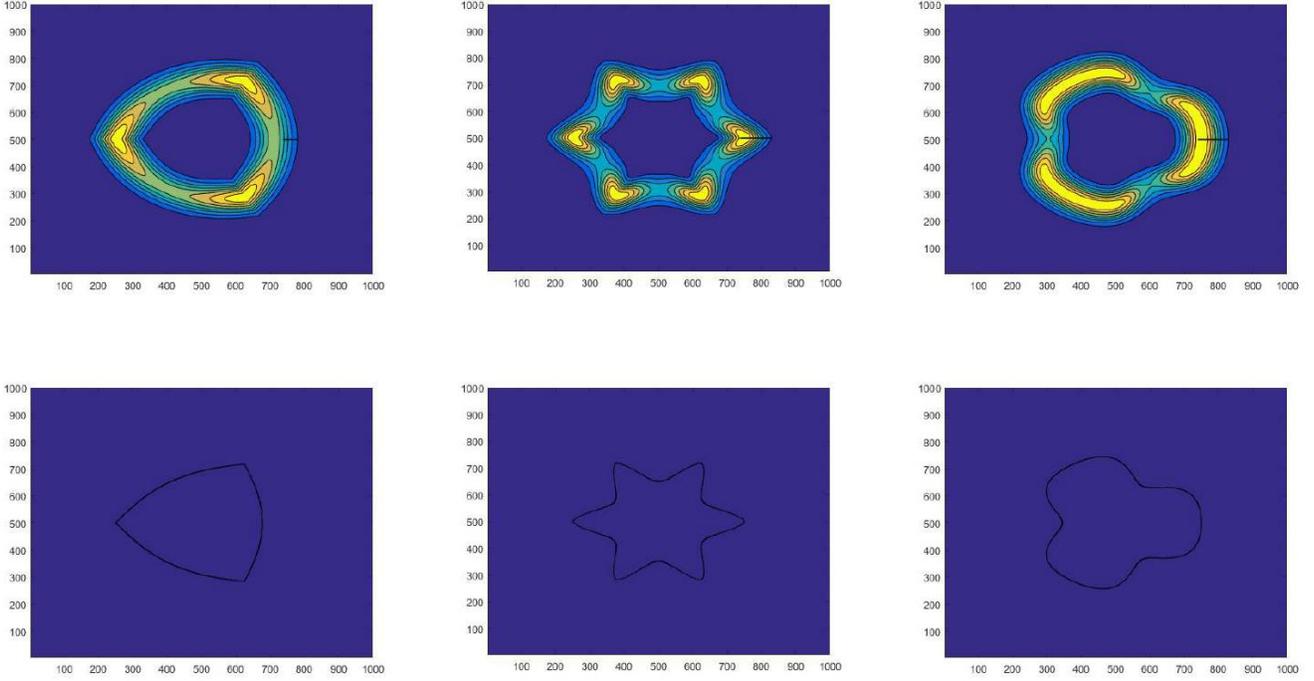

Fig. 6. (Top Row) The shape matching algorithm determines relative information content of shapes compared to the circle. The brightest contour shows regions of high information content that form 'dinstinctive features,' consisting of sharp and blunt edges/corners. (Bottom Row) Corresponding 'base shapes'.

Once in this state, a mean shape vector is calculated by taking the average of all of the corresponding points on all of the face vectors. This mean face vector is subtracted from each face, to reduce the images into their differences from the mean, a process known as normalizing or normalization. These normalized vectors are then processed using principal component analysis to create a series of eigenvectors that best describe the distribution of the shapes in the training set, referred to in the Turk and Pentland paper as "face space" [4]. Principal component analysis is used to find a smaller number of uncorrelated variables from a larger set of data.

One of the key components of this process is the reduction in the size of the data set needed to compare after the principal component analysis from an M·N eigenvalues and eigenvectors to a number of vectors that is less than the total number of input images. This can be done because when there are fewer input images than there are data points, there will be a smaller number of useful vectors whose corresponding eigenvalues are non-zero than the total number of data points included in the calculation. By using only the vectors with non-zero eigenvalues, the process can be conducted accurately with significantly fewer calculations.

The application of these eigenfaces to face or image recognition can be conducted after the generation of this base set. To do this, a new image is input to the system and put through the same process as the training images to reduce it to a normalized vector. This vector is then run through the following simple mathematical operation:

$$\omega_k = u_k^T(\Gamma - \Psi)$$

where, $\omega_k$, is the projection of the input face, $\Gamma$, into face space, $u_k$ is one of the generated eigenvectors, and $\Psi$ is the mean eigenface. The vector composed of the values of $\omega_k$ for all $k$ is known as $\Omega$. The $\Omega$ vectors can then be translated into a weighted score by calculating the Euclidian distance of the input face from a face class, $\Omega_k$. This is done using the equation:

$$\varepsilon_k = \|(\Omega - \Omega_k)\|$$

where $\varepsilon_k$ is the distance an input face is from a given face class, $\Omega_k$. Because of its effectiveness, for Turk and Pentland and the simplicity of implementation, this is the method of weighting and categorizing shapes used in this work. Once these weights are calculated, the best match is determined to be the image with the smallest value of $\varepsilon_k$ not exceeding a set value $\theta$ known as the cutoff criteria. Using this eigenface algorithm output is a vector called *outWeights* that describe the closeness of the input to each of the library shapes. Using the *outWeights*, it's possible to determine quality of matching using the following equation:

$$Mscore = \frac{1 - \min(w_{oi})}{\sum_i^n w_{oi}/n}$$

where *MScore* > $\gamma$ for it to be counted. This ensures the matching is done with a high enough threshold in the *MatchScore* (abbreviated *MScore*). In the following section, we apply the algorithm to a several test cases both to analyze the capabilities and limitations of the algorithm.

## IV. SIMULATION EXPERIMENTS

In this section, we apply the algorithm and evaluate it under ideal and non-ideal conditions. The shapes first compared are shown in Figure 7. These three shapes already exist in the library of interesting shapes used for our experiments and hence we would like to analyze how well the algorithm matches to the correct shape under ideal conditions.

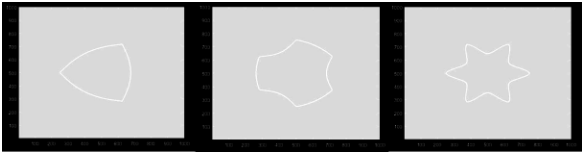

Fig. 7. Shapes used for ideal comparison.

Table I summarizes the results. As shown, all of the shapes are correctly matched as expected. With this elementary example, we then introduce various forms of distortion to the shapes and those modified shapes are shown in Fig. 8, 9 and 10. The parameters modified are shown in Table II, III and IV respectively.

TABLE I. MATCH PERFORMANCE FOR IDEAL SHAPES

| Shape | MScore | outWeight | Nearest Match | Average outWeight | Correct Match |
|---|---|---|---|---|---|
| Rounded Triangle | 1 | 0 | 2.1e-3 | 3.5e-3 | Yes |
| 3-Faced Blunt | 1 | 0 | 2.3e-3 | 3.9e-3 | Yes |
| 6-Pointed Star | 1 | 0 | 2.8e-3 | 4.9e-3 | Yes |

TABLE II. PARAMETERS FOR NEAR ROUNDED TRIANGLE

| Near Rounded Triangle | |
|---|---|
| Base | Modified |
| $a = b = 1$ | $a = b = 1$ |
| $m_1 = 3$ | $m_1 = 3$ |
| $m_2 = 3$ | $m_2 = 3$ |
| $n_1 = n_2 = n_3 = 1500$ | $n_1 = n_2 = n_3 = 1650$ |

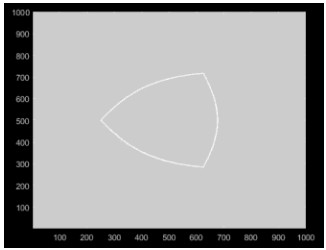

Fig. 8. Near rounded triangle

TABLE III. PARAMETERS FOR NEAR 3-FACED BLUNT

| Near 3-Faced Blunt | |
|---|---|
| Base | Modified |
| $a = b = 1$ | $a = b = 1$ |
| $m_1 = 6$ | $m_1 = 6$ |
| $m_2 = 6$ | $m_2 = 6$ |
| $n_1 = 60$ | $n_1 = 66$ |
| $n_2 = 55$ | $n_2 = 60.5$ |
| $n_3 = 10$ | $n_3 = 11$ |

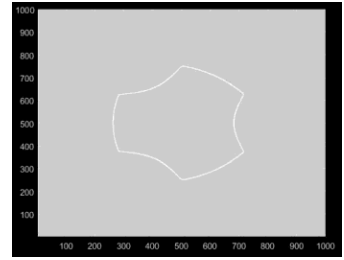

Fig. 9. Near 3-faced blunt.

TABLE IV. PARAMETERS FOR NEAR 6-POINTED STAR

| Near 6-Pointed Star | |
|---|---|
| Base | Modified |
| $a = b = 1$ | $a = b = 1$ |
| $m_1 = 6$ | $m_1 = 6$ |
| $m_2 = 6$ | $m_2 = 6$ |
| $n_1 = .2$ | $n_1 = .205$ |
| $n_2 = n_3 = 1.7$ | $n_2 = n_3 = 1.71$ |

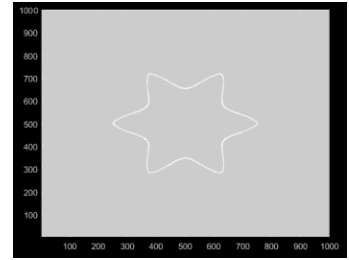

Fig. 10. Near 6-pointed star.

With these modifications, we find the algorithm can still correctly match to the correct shapes as shown in Table V. The slight decrease in Matchscore is expected especially for shapes with many features such as the six-pointed star. Any slight modifications to the shape parameters have larger impact than smoother, rounder shapes.

TABLE V. MATCH PERFORMANCE FOR SIMILLAR SHAPES

| Shape | MScore | outWeight | Nearest Match | Avg outWeight | Correct Match |
|---|---|---|---|---|---|
| Near Rounded Triangle | 1 | 8.9e-7 | 2.1e-3 | 3.5e-3 | Yes |
| Near 3-Faced Blunt | 1 | 1.1e-5 | 2.3e-3 | 3.9e-3 | Yes |
| Near 6-Pointed Star | 0.92 | 3.6e-4 | 2.6e-3 | 4.7e-3 | Yes |

### A. Effect of Sensor Error

Applying random error to the positions of the points used to describe a shape rapidly decreases the matching capability of the algorithm (see Fig. 11). Using this selected shape, a 7% error makes it difficult to conclusively distinguish the input shape from any of the shapes in the library. Beyond this level of error, the effects of the random percentage error on the matching capability of the algorithm are dominated by the randomness of the applied error.

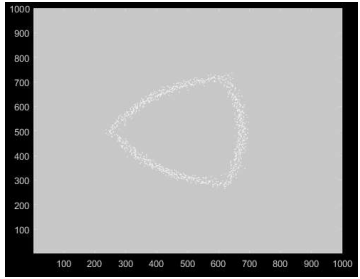

Fig. 11. Example of noise applied to the rounded triangle.

Fig. 12 shows the MatchScore against the applied standard error. These figures show that, as expected, increasing the error applied decreases the quality of the matches in terms of the MatchScore.

*B. Effect of Rotation*

The effect of rotation on MatchScore (abbreviated Mscore) is shown in Fig. 13. These results are somewhat better than expected. For all the included shapes the technique is resilient to slight rotations. Given that the deviation was small and almost identical for most of the shapes it is likely that this same trend would apply to most possible shapes. This resilience would allow for an approach to matching unknown shapes to the library regardless of their degree of rotation. This process would be simple and could be applied in a number of ways. Regardless of the specific manner of application the curves show a clear pattern. As the shape is rotated its MatchScore deviates from and then returns to 1. This means that for a series of smooth rotations the slope of the MatchScore curve could be used to identify the degree and direction of rotation necessary to match the shape accurately. Or, if computational power is not an issue, a full rotation on the gathered data could be performed and the angle producing a minimum MatchScore could be taken to be the best matching angle. The applicability of this approach confirms the earlier assumption that this technique will allow matching of rotated shapes. Specifics of this technique will be outlined in the discussion section.

*C. Effect of Changing Shape Parameters*

Slight modifications to the shape parameters have significant impact on the overall shape. Changing the *m* value of the shape changes the number of degrees of rotational symmetry it has (Fig. 14). These alterations would have been applied asymmetrically; however, doing so produces very strange and inconsistent shapes that differ so significantly from the base shape that attempting to match them in the same category as the base shape does not make sense.

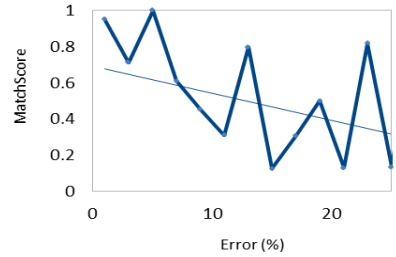

Fig. 12. Effect of normal error in points position to Matchscore presuming circle as ideal shape.

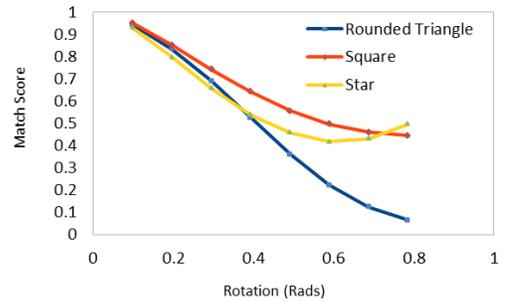

Fig. 13. Effect of Rotation on MScore for various base shapes.

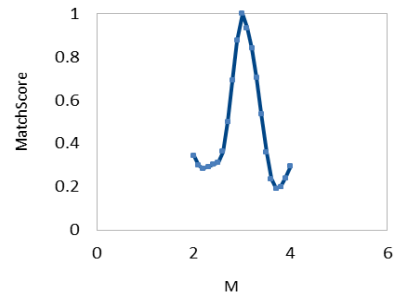

Fig. 14. Effect of varying *M* on MScore.

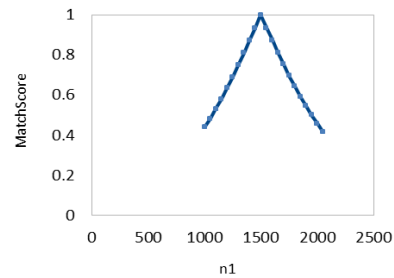

Fig. 15. Effect of varying $n_1$ on MScore.

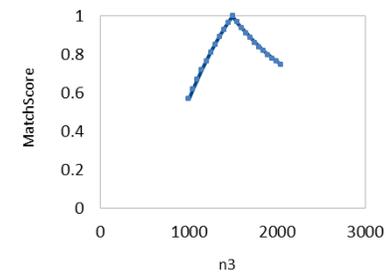

Fig. 16. Effect of varying $n_2=n_3$ on Mscore.

Altering $n_1$ adjusts the curvature of the shape (Fig. 15). Higher values produce straighter sides and lower values produce more rounded ones. Altering $n_2$ and $n_3$ symmetrically has a similar but more muted effect (Fig. 16). Altering them independently has the same affect but it is asymmetric. Finally, altering $a$ and $b$ symmetrically increases the size of the shape. This was not tested because it does not affect the shape only the size of the object. Altering them asymmetrically produces discontinuities and was thus not tested. Once a sample shape is correctly sized, then it requires just modification of the 5 shape parameters and rotation angle to obtain a suitable match. Pre-assembly of samples and categories can further speed-up the process of determining a match or not.

## V. Discussion

The shape matching approach presented here was shown to be accurate and resilient to a number of non-ideal conditions including error, rotation, and small changes in the shape of the object. The approach is shown to handle noise of up to 20% with a graduated degradation in performance. These are however early results. The tests were conducted on artificially generated shapes, rather than data collected using real equipment to extract location points. While the integration of the simulation and the technique used in that simulation does demonstrate that this technique is capable of extracting these values and matching them successfully.

Currently the method does not ascribe a value to recognized objects based on observations of the object, but requires the robot to call one that is preassigned. However as we suggest in this work, the pre-assigning of interesting shapes can be done automatically, through extrapolation from families of interesting shapes using the Superformula. In addition to simply comparing all objects to a predefined library of possible shapes, shapes could be compared both to that library and a library created during the mission of observed shapes. Any detected obstruction that is processed through the matching component could be categorized and then added to the secondary library with an associated category after processing. New objects could then be compared against these to determine if it is significantly different from the previously observed objects. Such a metric of difference could serve as a quantification of the novelty as it would constitute a metric of how different a detected object is from other observed objects. This approach would be easy to implement, but determining the proper method of weighting these results would be a significant task.

## VI. Conclusions

In this work, we propose a new approach to automating the process of identifying and categorizing interesting shaped-objects. This approach can enable automated geological exploration, enabling robots to find rock samples and objects of interest through their distinctive shape. The method uses a Gaussian model to determine net information content of two-dimensional objects in comparison to a circle. An extension of the algorithm for use on three-dimensional objects is proposed. The information content identifies the relative 'importance' of features of a shape automatically. Using this approach, a shape sample is compared against a library of shapes using the eigenfaces approach. The technique is successfully shown to match sample objects to a library. An increase in sensor noise is shown to result in a graduated decrease in match performance. For practical applications, it is found that the shapes may need to be analyzed in terms of multiple rotation angles to maximize match. The general robustness to noise and shape parameters shows that the method is suitable for use with real sensors, including vision cameras and laser-ranger finders. Work is underway in testing the algorithm with real rock samples and meteorites.

## Acknowledgment

The authors would like to gratefully acknowledge Dr. James Middleton and Dr. Spring Berman for helpful discussions. The authors would like to thank Dr. Laurence Garvey from ASU's Center for Meteorite Studies for his input into this research.

## References

[1] National Research Council. 2011. Vision and Voyages for Planetary Science in the Decade 2013-2022. Washington, DC: The National Academies Press.
[2] C.B. Agee, N.V. Wilson, F.M. McCubbin, et al., "Unique Meteorite from Early Amazonian Mars". Science. 339, 2013, pp. 780–785.
[3] M.J. Genge, J. Larsen, M. Van Ginneken and M.D. Suttle, "An Urban Collection of Modern-Day Large Micrometeorites," *Geology*, 2016,
[4] M. A. Turk and A. P. Pentland, "Face recognition using eigenfaces," *IEEE Computer Vision and Pattern Recognition*, 1991, pp. 586-591.
[5] J. Gielis, "A Generic Geometric Transformation That Unifies A Wide Range of Natural and Abstract Shapes." American Journal of Botany 90.3, 2003, pp. 333-338.
[6] B. Charrow, et al. "Information-Theoretic Planning with Trajectory Optimization for Dense 3D Mapping." *RSS, Rome,* 2015.
[7] D. Rao, A. Bender, S. B. Williams and O. Pizarro, "Multimodal information-theoretic measures for autonomous exploration," *IEEE ICRA*, 2016, pp. 4230-4237.
[8] E. Kaufman, T. Lee and Z. Ai, "Autonomous exploration by expected information gain from probabilistic occupancy grid mapping," *IEEE SIMPAR*, 2016, pp. 246-251.
[9] S. J. Moorehead, R. Simmons and W. L. Whittaker, "Autonomous exploration using multiple sources of information," *IEEE ICRA*, 2001, pp. 3098-3103.
[10] S. Belongie, J. Malik and J. Puzicha, "Shape matching and object recognition using shape contexts," in *IEEE Transactions on Pattern Analysis and Machine Intelligence*, vol. 24, no. 4, 2002, pp. 509-522.
[11] H. Su, S. Maji, E. Kalogerakis, E. Learned-Miller, "Multi-view Convutional Neural Networks for 3D Shape Recognition," International Conference on Computer Vision, pp. 1-9.
[12] J. Thangavelautham, G.M.T. D'Eleuterio, "Tackling Learning Intractability through Topological Organization and Regulation of Cortical Networks," IEEE Transactions on Neural Networks and Learning Systems, Vol. 23, No. 4, pp. 552-564, 2012.
[13] J. Thangavelautham, P. Grouchy, G.M.T. D'Eleuterio, Application of Coarse-Coding Techniques for Evolvable Multirobot Controllers, To Appear in Chapter 16, Computational Intelligence in Optimization-Applications and Implementations, Vol. 7, Springer-Verlag, Berlin, Germany, 2010, pp. 381–412.
[14] J. Thangavelautham, G.M.T. D'Eleuterio, "A Neuroevolutionary Approach to Emergent Task Decomposition," Lecture Notes in Computer Science 3242:991-1000.